 \documentclass[cameraready]{Interspeech}
\title{TMASC: Transmasculine Attitude and Speech Corpus}

\author[affiliation={1,2}, orcid=0000-0002-8483-0540, equalcontribution, correspondingauthor]{Sidney}{Wong}

\address{
    $^1$ Centre for Sustainability Research, University of Otago, New Zealand \\
    $^2$ Te Pūnaha Matatini Centre of Research Excellence for Complex Systems, New Zealand
}

\email{sidney.wong@otago.ac.nz}


\usepackage{comment}

\begin{document}

\maketitle


\begin{abstract}
    We introduce the Transmasculine Attitudes and Speech Corpus (TMASC), a multimodal corpus of 196 transmasculine individuals, including questionnaire responses and 66 audio recordings. The questionnaire includes items exploring the vocal health of transmasculine individuals. The audio recordings include cough and throat-clearing samples, a reading passage, and additional session-specific questions. This paper outlines the development of this corpus and the data collection procedures. To illustrate the utility of this corpus, we present three case studies demonstrating how this crowd-sourced multimodal corpus can be used to support transmasculine individuals. These include the integration of perceptual and acoustic data, the identification of group-level characteristics, and the calibration of acoustic measurements.
\end{abstract}

\section{Introduction}

    A gender-congruent voice predicts greater well-being for transmasculine individuals \cite{watt_masculine_2018}. However, research into the vocal needs of transmasculine individuals has not received the same attention as that of transfeminine individuals. This imbalance has stemmed from the prevailing belief that testosterone replacement therapy (TRT) sufficiently lowers pitch, or the fundamental frequency ($f_0$), to achieve a satisfactory voice \cite{van_borsel_voice_2000}\cite{tsjoen_impact_2006}. In fact, the needs of these groups are vastly different. In addition to TRT \cite{deuster_voice_2016}, the use of chest binding \cite{peitzmeier_health_2017}, top surgery \cite{lo_russo_masculine_2017}, glottoplasty \cite{meister_perceptual_2017}, and informal vocal practices such as smoking to lower speaking pitch \cite{azul_transmasculine_2017} all have an impact on the vocal health of transmasculine individuals. While vocal training provided by speech and language therapists often works with transmasculine patients to achieve a masculine speaking pitch based on cisgender population norms, speaking $f_0$ alone does not equate to a gender-affirming voice \cite{mcneill_perception_2008}.
    
    In this paper, we introduce the Transmasculine Attitudes and Speech Corpus (TMASC). The TMASC corpus was originally developed as part of the author's Master's thesis \cite{wong_exploring_2017}. The primary aim of this crowd-sourced multimodal corpus is to provide researchers, health practitioners, and community members with access to a high-quality data source that can be used to investigate the vocal needs of transmasculine individuals. The corpus consists of a questionnaire addressing the psychosocial and communicative effects that may impact this diverse population. Optionally, participants were encouraged to provide a speech sample (including a cough and throat-clearing sample, as well as a reading passage), which could be analysed alongside their questionnaire responses. Therefore, TMASC offers a unique view of the vocal needs of transmasculine individuals at a community level. In order to demonstrate the benefits of our multimodal approach, we consider how acoustic information can be used to determine population-level socio-acoustic norms.

\section{Related Works}

    An individual's voice is frequently used as an external cue of gender \cite{zimman_transgender_2018}. This is particularly apparent for some transmasculine individuals who have experienced changes to their voice during TRT \cite{irwig_effects_2017}\cite{hancock_trans_2017}. Physiological changes to the speech apparatus as a result of TRT are only one aspect of the self-perception of gender. In reality, multiple factors impact the vocal situation of transmasculine individuals. These factors can be grouped into: \textit{presentational factors}, including the method or methods used to change vocal gender presentation, anatomical dimensions of vocal organs, and gender-related vocal features; \textit{attributional factors}, including self-perception of voice in regard to gender as well as gender attribution by others; \textit{normative factors}, including standards of femininity and masculinity; and lastly, \textit{diversity}, involving the subjective positioning of gender, and method(s) used to change gender presentation \cite{azul_transmasculine_2017}.
    
    With this in mind, a holistic approach is therefore required to understand the vocal health needs of transmasculine individuals. Health practitioners would use diagnostic questionnaires to determine the vocal needs of transgender individuals. Some of the earliest questionnaires included the Voice Handicap Index (VHI) \cite{jacobson_voice_1997}, the Transgender Self-Evaluation Questionnaire (TSEQ) \cite{dacakis_development_2013}, and the Transgender Voice Questionnaire for Female-to-Male (TVQ-FTM) \cite{dacakis_exploring_2016}. Once again, it is important to note that these diagnostic tools were developed to address the needs of transfeminine individuals. Only recently have we seen the development of the Transgender Voice Questionnaire for Female-to-Male (TVQ-FTM), responding to the unique needs of transmasculine people. One limitation of diagnostic questionnaires is that the results cannot be easily interpreted alongside acoustic information. Furthermore, these diagnostic tools are used to determine the needs of individuals and not at a community level.

    As research into the vocal needs of transgender individuals has moved from a clinical to a social domain, we now know that a gender-congruent voice does not necessarily align with cisgender population benchmarks \cite{merritt_toward_2020}. Therefore, there is an increased need to establish community-appropriate benchmarks outside a diagnostic setting \cite{merritt_perceptual_2020}\cite{merritt_revisiting_2022}. With the popularisation of personal electronic devices (such as smartphones and laptops), crowd-sourcing is becoming an increasingly cost-effective approach in the development of corpora for clinical purposes \cite{bunnell_modeltalker_2017}. For example, a \textit{Palette of Transmasculine Voices} included the voices of 20 American English-speaking transmasculine men, transmasculine non-binary people, and cisgender men \cite{dolquist_clinical_2024}. The corpus included Consensus Auditory-Perceptual Evaluation of Voice sentences, the \textit{Rainbow Passage}, and a novel set of sentences. While the corpus is restricted to a limited sociolinguistic context, the results show that it is possible to establish community-appropriate population benchmarks at scale.
    
\section{Methodology}

    Influenced by existing work in dialectology, we adopted a crowd-sourcing approach in the development of TMASC. As mentioned previously, the corpus consisted of a questionnaire and an optional speech sample. Of the 196 completed questionnaires, 66 participants provided a speech sample through their personal electronic devices. The participants were recruited online through email and social media networks. The data were collected over a three-month period between July 2017 and October 2017. In this section, we provide an overview of our methodology, including the ethical considerations, participants, data collection procedures, and data availability.

\subsection{Ethical Considerations}

    TMASC was developed in consultation with researchers and community members. All participants were provided with information detailing the purpose and aim of the study. We received approval from the University of Canterbury Human Ethics Committee (HEC 2017/12). Only those who were 18 years or older were included in our analysis. All raw, unedited data were stored for a period of five years. The contact details of participants were stored separately from the speech and questionnaire data. Following the questionnaire, participants were presented with a debriefing screen, and links were provided to additional resources and further research into transgender and gender non-conforming individuals. If participants desired more information about vocal health and surgical procedures for voice masculinisation, such as thyroplasty type 3, this was also provided. A summary of results was sent to participants on request.

\subsection{Participants}

    The typical profile of a participant was aged between 22 and 37.5; was of European descent and resided in a predominantly English- or German-speaking country; did not smoke; identified as queer and was equally attracted to female- and male-bodied people; was assigned female at birth and preferred `he/him/his' pronouns (for the English-speaking participants); had disclosed their gender history to their family, partner, and friends; and, at the time of data collection, was binding and was experiencing negative health impacts from their binding techniques.

\subsubsection{Geographic Source}

    In terms of geographic spread, participants spent their formative years in English-speaking countries such as the United States ($n=55$), Australia ($n=23$), New Zealand ($n=14$), Canada ($n=11$), and the United Kingdom ($n=11$), and predominantly German-speaking countries such as Germany ($n=31$) and Switzerland ($n=10$). This broad geo-linguistic distribution of the corpus can be observed in the language conditions of the reading passages, where the majority were in English ($n=50$) and the remaining recordings were in German ($n=16$).

\subsubsection{Gender}

    Participants were asked to input words that best described their gender identity (\textit{What word(s) best describe your gender identity?}). Many participants included multiple terms to describe their gender identity. An exploratory textual analysis was conducted to identify the most significant gender identities in the study. Participants used the terms \textit{male}, \textit{man}, \textit{non-binary}, \textit{transmasculine}, \textit{trans}, \textit{transman}, or \textit{transgender}, as well as \textit{genderqueer} or \textit{agender}, to describe their gender. Eight participants specified \textit{female-to-male} or \textit{FtM}.
    
\subsection{Data Collection}

    In this section, we describe the data collection procedures, including the software, as well as the speech sample and questionnaire design.

\subsubsection{Software}

    The questionnaire results and speech samples were collected and housed on the \textit{Language, Brain and Behaviour Corpus Analysis Tool} (LaBB-CAT; \cite{fromont_labb-cat_2012}), which is a browser-based corpus analysis tool. Participants had the option to complete the study using their own personal electronic devices (e.g. laptops, desktop computers, tablets, mobile phones, etc.). Due to the crowd-sourced nature of the recordings, sampling rates and recording conditions varied across participants' devices.

\subsubsection{Speech Sample}

    Participants were given the choice to provide speech samples. The speech sample comprised four components, including a throat-clearing sample, a cough sample, and the reading passage \textit{North Wind and the Sun} in a language of their choosing. Translations were retrieved from the Aesop Language Bank (including English, German, Dutch, French, Russian, Hindi, Traditional Chinese, Simplified Chinese, Japanese, and Korean) and were proofread for clarity and consistency. At the conclusion of the speech sample, we included additional questions for participants to indicate whether they were binding during the session. All sound files were saved in .wav format.

\subsubsection{Questionnaire}
    
    The questionnaire consisted of 60 questions, grouped into relevant blocks to address different vocal needs of transmasculine individuals \cite{azul_transmasculine_2017}. The questions included a mix of binary and multivariate responses. Open-ended questions were limited; however, participants were able to answer in relation to their gender or sexual orientation in a free text box. Questions one to four in the first block addressed self-perception. This block included questions related to participants’ vocal satisfaction and how they used their voice. Questions 5 to 32 were related to participants’ vocal and communicative factors. Questions 33 to 40 were only available to participants if they indicated they had taken testosterone. The questions in this block included testosterone usage history and how testosterone had affected the participants' voice. If participants indicated they did not use testosterone, they were redirected to the penultimate block of questions, which included questions about vocal intervention methods in questions 41 to 48. Questions 49 to 59 were included to gather demographic information about participants, such as age and country of origin, as well as sensitive questions regarding their gender identity, assigned sex at birth, and sexual orientation.

\subsection{Data Availability}

    The anonymised questionnaires and acoustic data can be accessed through the project’s Open Science Foundation (OSF) repository\footnote{\url{https://osf.io/tg8bc/}}. While we envisioned the corpus to be freely available, user registration and password protection were enabled to keep track of the research outputs from the corpus.

\section{Case Studies}

    In lieu of a results section, we describe the following use cases to demonstrate how TMASC can be used for socio-acoustic and perceptual research. We used R through RStudio for data analysis and data visualisation. Acoustic measurements (such as mean and mode $f_0$) were retrieved through the Praat integration in LaBB-CAT and Robust Epoch And Pitch EstimatoR (REAPER) \cite{talkin_reaper_2015}. Our three case studies show how the questionnaire data can be used to understand self-perceived vocal health, establish community-level benchmarks, and how the speech samples can be used to calibrate community-specific acoustic measures. 

\subsection{Self-Perceived Vocal Health}

    \begin{figure}[h]
      \centering
      \includegraphics[width=\linewidth]{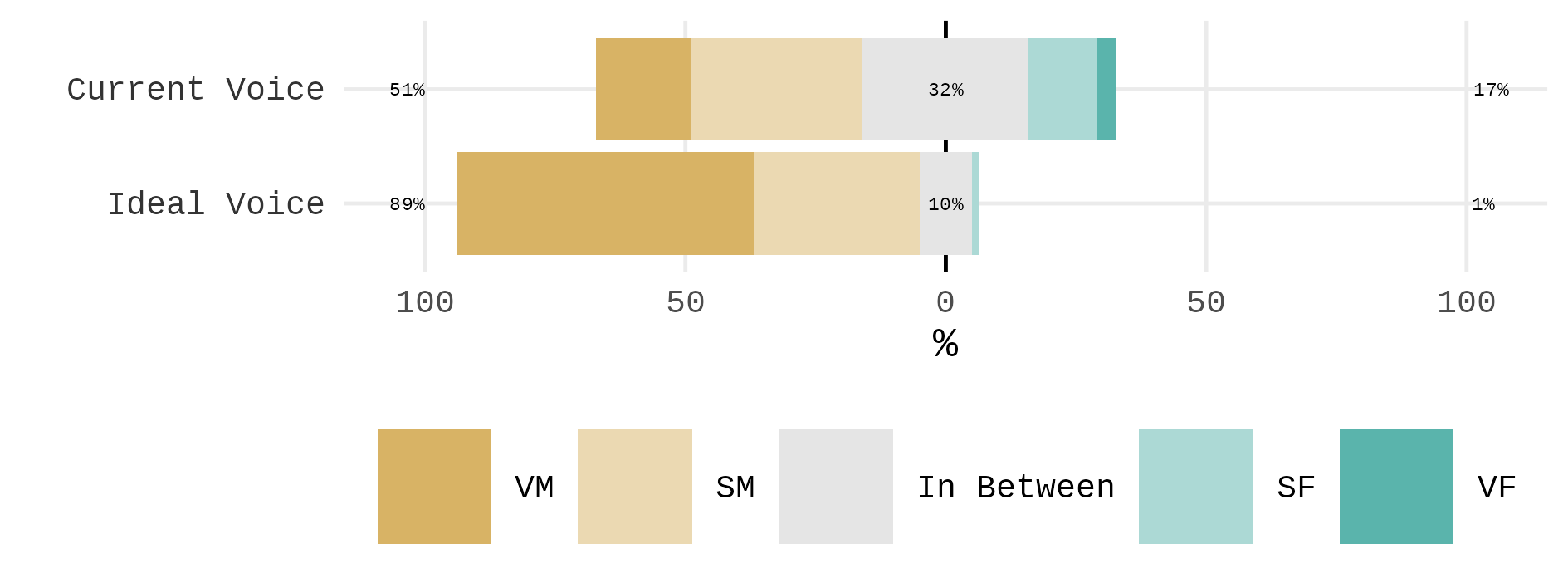}
      \caption{Stacked bar graph of perceived vocal satisfaction.}
      \label{fig:stacked_self_perceived}
    \end{figure}

    \begin{figure}[h]
      \centering
      \includegraphics[width=\linewidth]{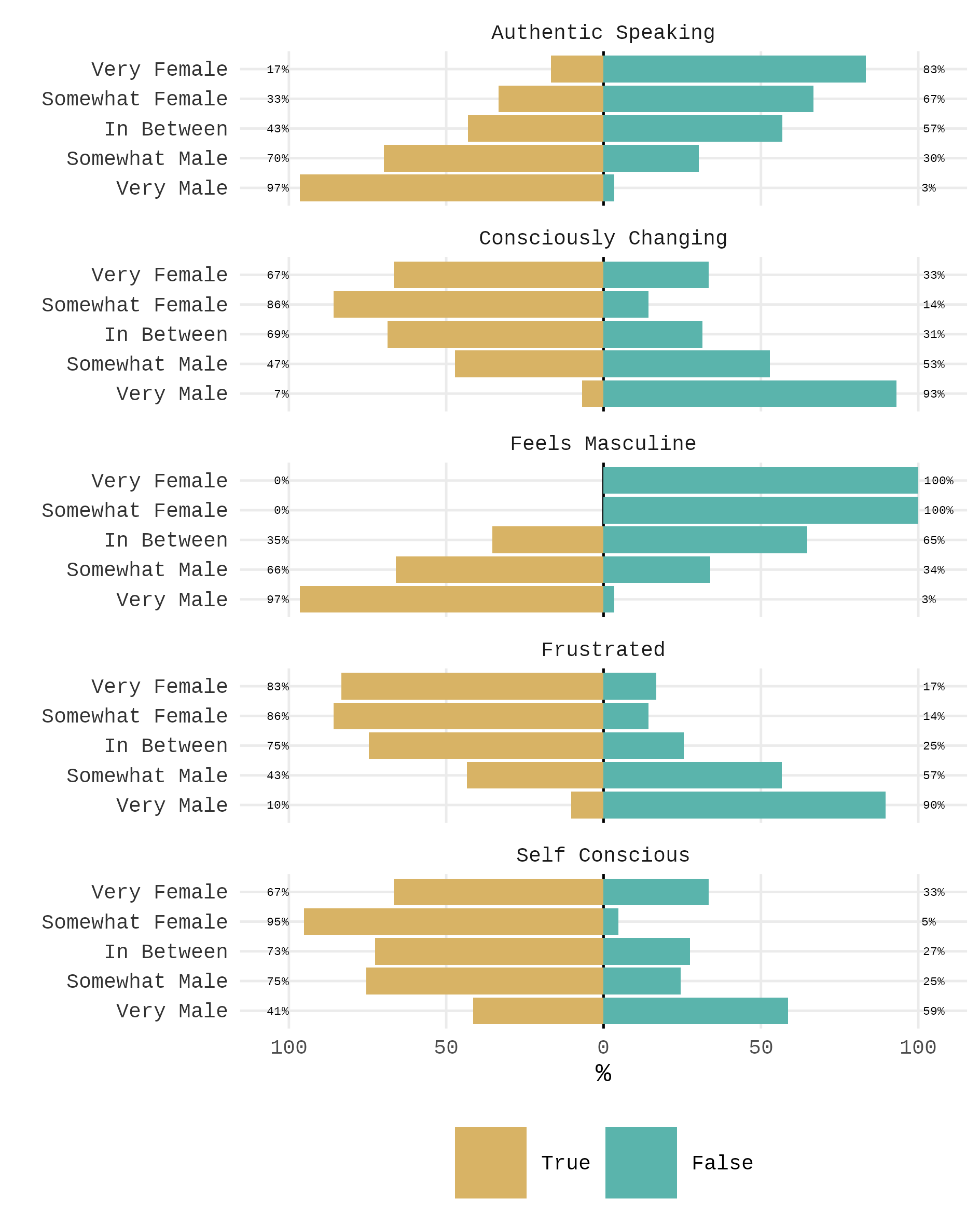}
      \caption{Stacked bar graphs of perceived limiting factors.}
      \label{fig:stacked_personal}
    \end{figure}

    The 196 completed questionnaires in the corpus offer a holistic view of the self-perceived vocal health of transmasculine individuals. To illustrate this, we present the results from self-perceived vocal masculinity (\textit{I believe my current voice is ...}) and ideal voice (\textit{My ideal voice is ...}), both measured on a five-point Likert scale, as shown in Figure \ref{fig:stacked_self_perceived}. Furthermore, we can combine questions to determine interactions, such as those between self-perceived vocal masculinity and personal impacts, as shown in Figure \ref{fig:stacked_personal}. The personal impacts include: \textit{I have a speaking voice that feels authentic to me} (Authentic Speaking), \textit{I'm consciously trying to change my voice} (Consciously Changing), \textit{my voice makes me feel masculine} (Feels Masculine), \textit{my voice frustrates me} (Frustrated), and \textit{I feel self-conscious about how strangers perceive my voice} (self Conscious).
    
\subsection{Community-Level Benchmarks}

\begin{figure}[h]
      \centering
      \includegraphics[width=\linewidth]{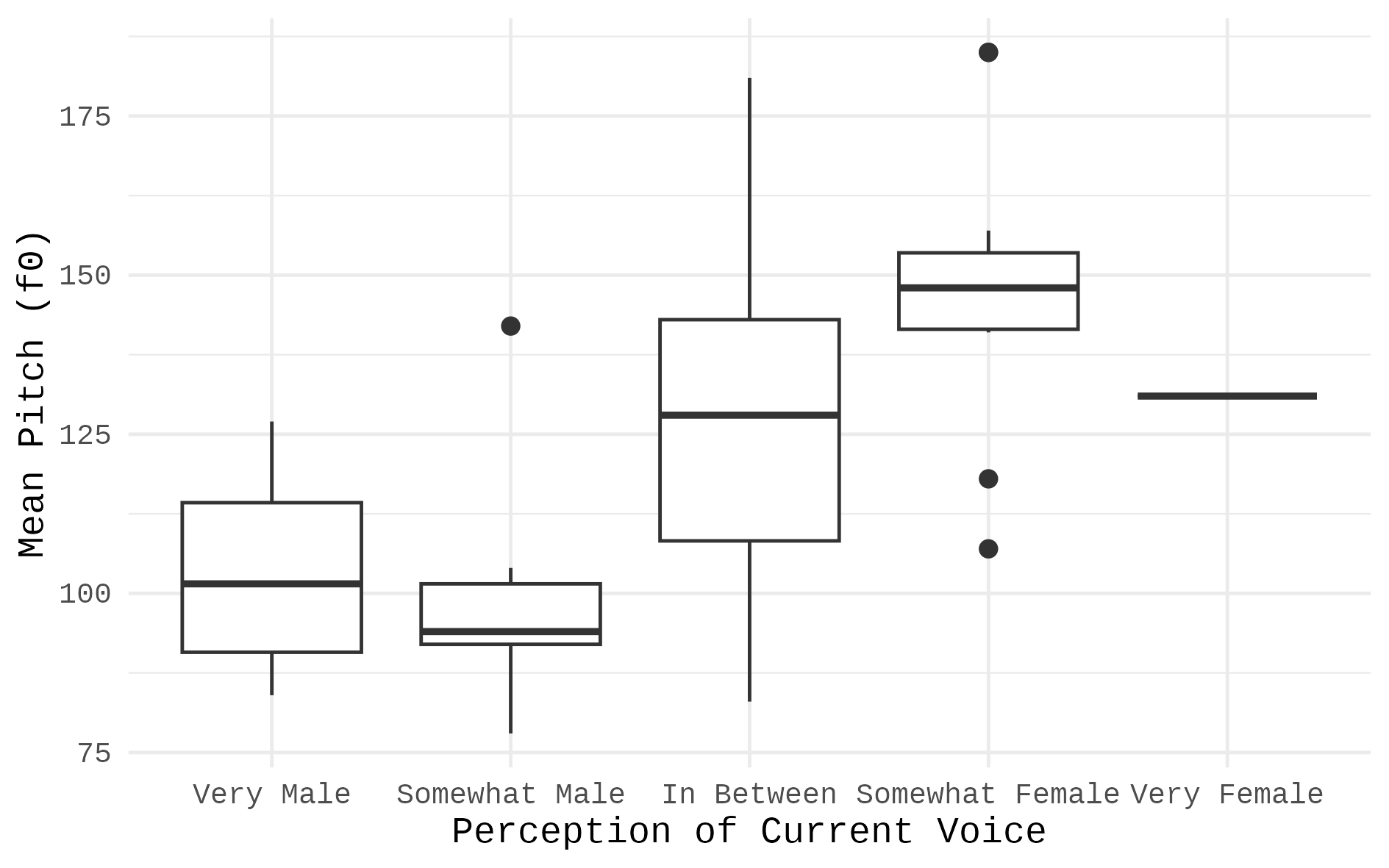}
      \caption{Boxplot of mean $f_0$ grouped by vocal masculinity.}
      \label{fig:self_perceived}
    \end{figure}

\begin{figure}[h]
      \centering
      \includegraphics[width=\linewidth]{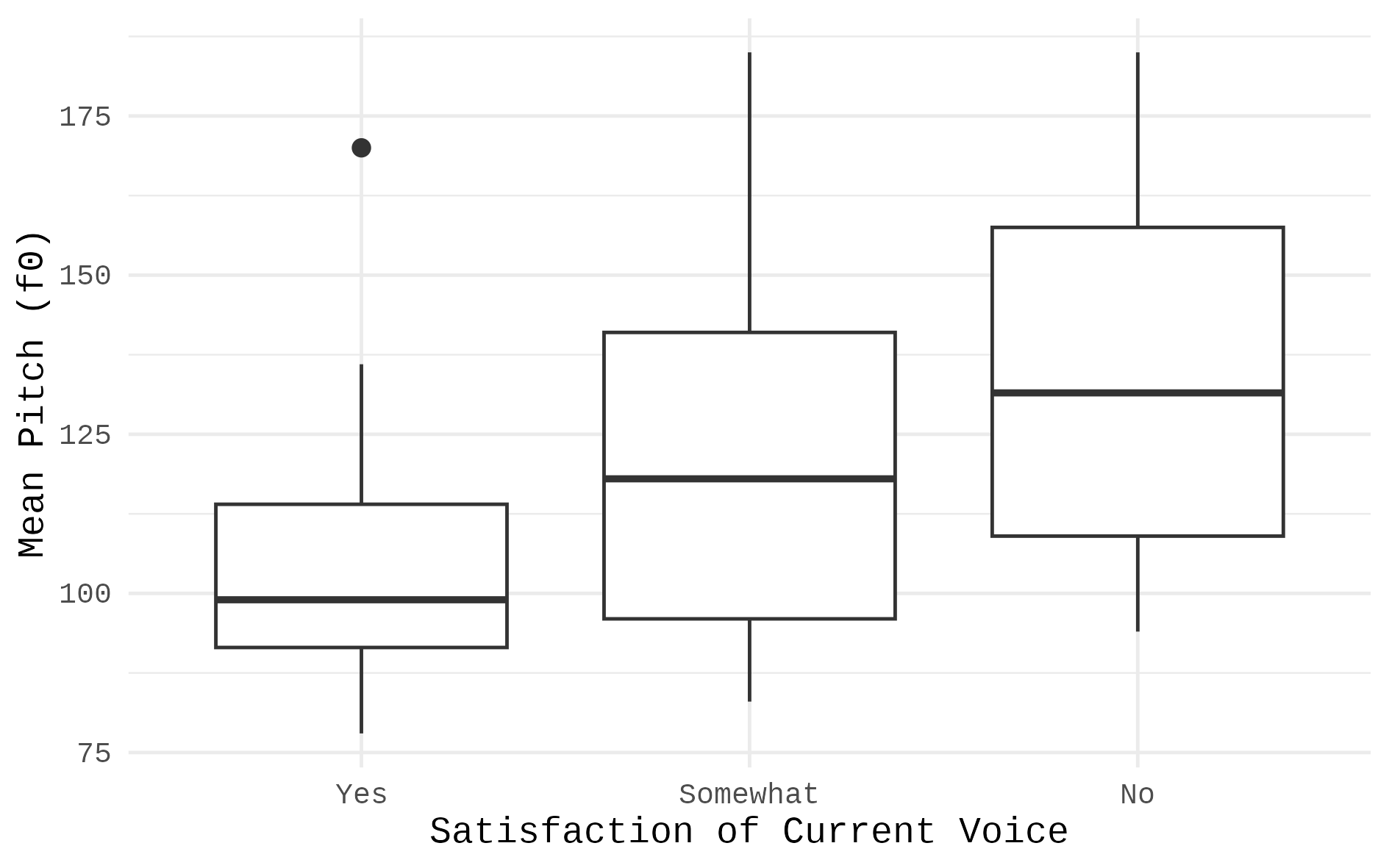}
      \caption{Density plot of mean $f_0$ grouped by vocal perception.}
      \label{fig:vocal_satisfaction}
    \end{figure}

    The key strength of the TMASC corpus is the inclusion of perceptual and acoustic information. The speech samples offer an additional dimension of analysis. We can use this self-report to determine perceived social correlates of speech in the case of mean $f_0$. As shown in Figure \ref{fig:self_perceived}, we visualised mean $f_0$ grouped by self-perceived vocal masculinity as box plots. We found that the majority of participants viewed their voice as in between; meanwhile, those who perceived their voice as somewhat male had a lower mean $f_0$ than those who perceived their voice as very male. A similar effect was observed for those who perceived their voice as somewhat female. This suggests there was only a weak linear relationship between self-perceived vocal masculinity and acoustic measurements, such as mean $f_0$. When we applied a similar analysis to vocal satisfaction (Are you satisfied with your present voice?), as shown in Figure \ref{fig:vocal_satisfaction}, we observed a more linear relationship between vocal satisfaction and mean $f_0$.

\begin{figure}[h]
      \centering
      \includegraphics[width=\linewidth]{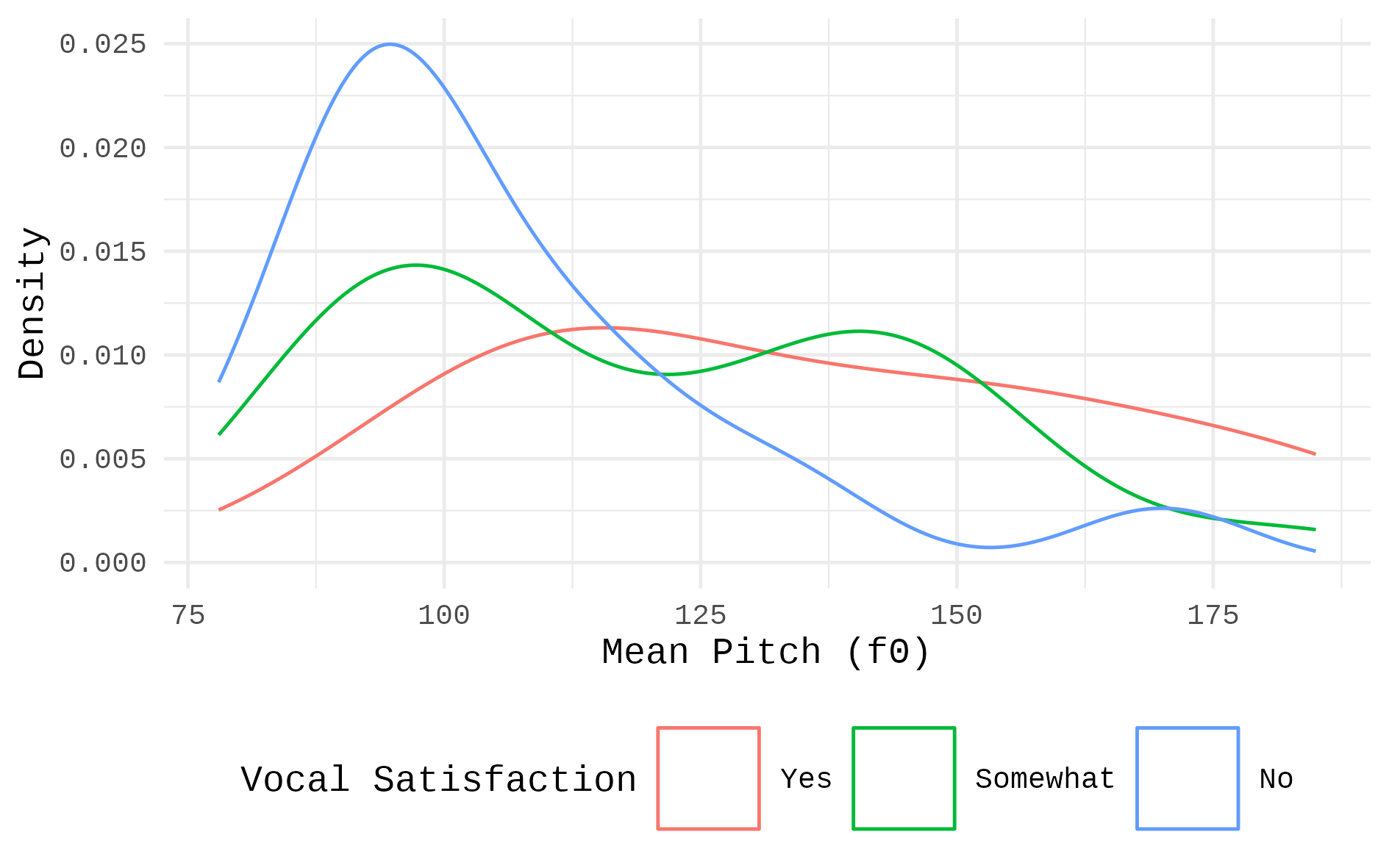}
      \caption{Density plot of mean $f_0$ grouped by vocal satisfaction.}
      \label{fig:vocal_satisfaction_mean}
    \end{figure}

    With access to the speech samples, it is also possible to interrogate the combined perceptual and acoustic information. For example, we visualised the same data used to create Figure \ref{fig:vocal_satisfaction} as a density plot. With reference to Figure \ref{fig:vocal_satisfaction_mean}, where we visualised mean $f_0$ grouped by vocal satisfaction, we observed a unimodal distribution for respondents who were satisfied with their voice (`yes'), with the peak below 100 Hz. However, for those who were somewhat satisfied with their voice (`somewhat'), we observed a bimodal distribution with the first peak around 100 Hz and the second peak around 130 Hz.

\subsection{Calibration of Acoustic Measures}

    We now show how the TMASC corpus can be used to calibrate acoustic measures used to determine a gender-conforming speaking voice. In terms of open-source software, Praat and REAPER are frequently used to retrieve mean $f_0$. Differences in pitch-tracking algorithms and estimation strategies may lead to systematic variation in the resulting acoustic measurements across tools. We use time on testosterone to calibrate and compare the acoustic measures retrieved from both Praat and REAPER. REAPER has been proposed as an alternative pitch-tracking tool which can accurately estimate glottal closure instants, voicing state, and $f_0$. Praat estimates mean $f_0$ measurements over the phrase, while REAPER estimates $f_0$ at each glottal closure. For example, the median $f_0$ measurement extracted from Praat via LaBB-CAT for the sample population was 137.8 Hz, and the mean was 150.7 Hz (range = 88.2–489 Hz). The mean $f_0$ measurement extracted from REAPER was 114 Hz, and the mean $f_0$ was 119.1 Hz (range = 78–185 Hz). The mean $f_0$ measurements extracted from REAPER were significantly lower than those extracted from Praat.

    \begin{figure}[h]
      \centering
      \includegraphics[width=\linewidth]{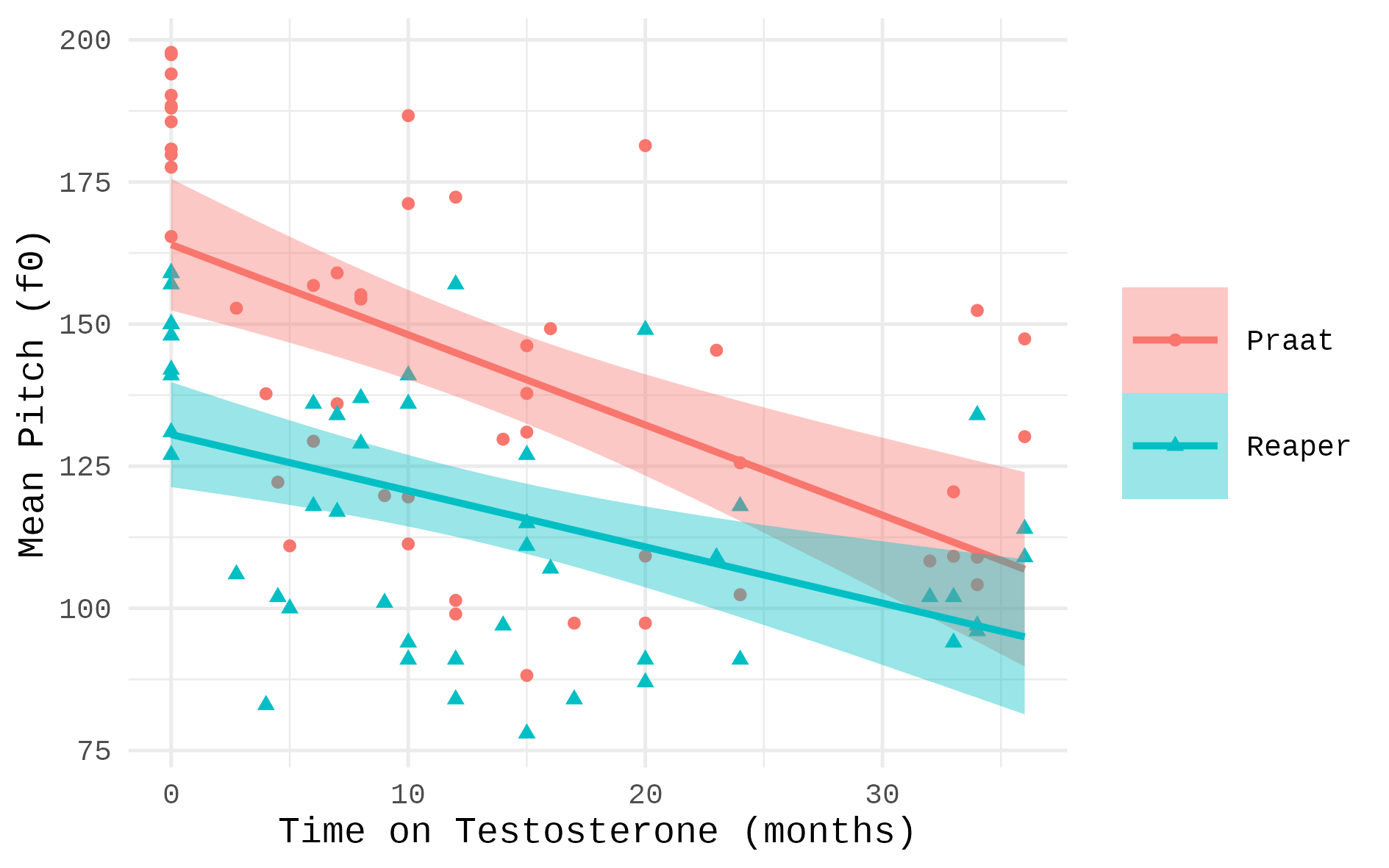}
      \caption{Scatterplot comparing mean $f_0$ measures by time on testosterone.}
      \label{fig:praat_reaper}
    \end{figure}

\section{Discussion \& Conclusion}

    The purpose of the Transmasculine Attitudes and Speech Corpus (TMASC) was not to develop a diagnostic tool for vocal health practitioners; instead, the corpus provides a valuable source of perceptual and acoustic information to understand the vocal needs of transmasculine individuals. More importantly, we showed that a crowd-sourcing approach is a cost-effective way to produce a corpus to establish community-appropriate benchmarks which would otherwise be missed in a clinical environment. We now address some of the limitations of TMASC. There are several limitations to the current study. 
    
    Firstly, the corpus is not a longitudinal study and does not include data regarding the individual effects of testosterone on transmasculine individuals. It is a cross-sectional demographic survey that attempts to explore vocal satisfaction and quality of life of transmasculine individuals abroad. The longitudinal assumptions made in the current study only explore population trends and do not account for changes occurring at the individual level. However, the results from the current study do suggest that systemic changes are needed in the management of transmasculine vocal health. 
    
    Secondly, the data were not collected under laboratory conditions; therefore, the acoustic measurements extracted may be severely impacted by extraneous factors not accounted for in the current study. These extraneous factors could include faulty equipment or background noise, which may affect the quality of the recording. Participants may also be influenced or distracted by their immediate surroundings, as the study was conducted online. $f_0$, as an acoustic measure, is relatively stable, and REAPER is sensitive to these signals. Furthermore, participants may feel more at ease providing a speech sample in a familiar environment, in contrast to laboratory conditions. However, validation studies should be conducted in the future to compare the acoustic quality of personal recording equipment and speech recorded under laboratory conditions.
    
    Thirdly, there is no control over who participates in a study that is conducted online, and there are no guarantees as to how many people will participate. Nearly all the participants who provided a speech sample came from English-speaking countries, and some from German-speaking countries, even though transcripts were available in a number of languages. However, the current study was able to reach over 185 participants in more than 15 different countries. The ability to collect data from so many transmasculine individuals across the globe suggests that the lack of diversity in the current study can be mitigated in the future with proper promotion and marketing. Furthermore, the fact that the questionnaire was in English may exclude those who do not speak English as an additional language.

\section{Acknowledgements}

    I would like to thank Vica Papp and Tika Ormond for their supervision. I would also like to thank Robert Fromont for designing the underlying architecture of TMASC. I would also like to thank the panel of researchers and community members for their input into the design of TMASC. This research was made possible through the UC College of Arts Master of Linguistics thesis scholarship and funding from the School of Language, Social and Political Science. Lastly, I would like to acknowledge Te Pūnaha Matatini, New Zealand's Centre of Research Excellence in Complex Systems, for enabling this work to be published through my Postdoctoral Fellowship.

\section{Use of Generative AI Disclosure}

    The author acknowledges the use of generative AI (Microsoft Copilot, GPT-5) for editing and polishing the manuscript following peer review to improve clarity, grammar, and consistency with British English orthography. No generative AI tools were used to produce substantive content.

\bibliographystyle{IEEEtran}
\bibliography{references}

\end{document}